\documentclass[lettersize,journal]{IEEEtran}

\IEEEoverridecommandlockouts                              % This command is only needed if 
\providecommand{\keywords}[1]
{
	\small	
	\textbf{Keywords:} #1A
}

\usepackage{amsfonts,amssymb}
\usepackage[caption=false,font=footnotesize]{subfig}
\usepackage[dvipsnames]{xcolor}
\usepackage{multirow}
\usepackage{multicol}
\usepackage{graphicx}
\usepackage{subfloat}
\usepackage{amsmath}
\usepackage{amssymb}
\usepackage{booktabs}

\usepackage{multirow}
\usepackage{colortbl}
\definecolor{tabtitle}{gray}{.8}
\usepackage{bm}
\usepackage{pifont}
\newcommand{\yes}{\text{\ding{51}}}

\usepackage{marvosym}
\usepackage{hyperref}

\newcommand{\eg}{\textit{e}.\textit{g}.}
\newcommand{\etal}{\textit{et al}.}
\newcommand{\ie}{\textit{i}.\textit{e}.}

\newcommand{\wrt}{\textit{w.r.t.}}

\title{\LARGE \bf
eMoE-Tracker: Environmental MoE-based Transformer for Robust Event-guided Object Tracking
}

\author{Yucheng Chen$^{1}$ and Lin Wang$^{ 2*}$% <-this % stops a space
\thanks{*Corresponding author}% <-this % stops a space
\thanks{$^{1}$Yucheng Chen is with the AI Thrust, The Hong Kong University of Science and Technology (Guangzhou), Guangdong, China.
        {\tt\small yuchengc0221@gmail.com}}%
\thanks{$^{2}$Lin Wang is with the School of Electrical and Electronic Engineering (EEE), Nanyang Technological University (NTU), Singapore, Email:
        {\tt\small alwang.ntu@gmail.com}}%
        }

\begin{document}

\maketitle
\thispagestyle{empty}
\pagestyle{empty}

%%%%%%%%%%%%%%%%%%%%%%%%%%%%%%%%%%%%%%%%%%%%%%%%%%%%%%%%%%%%%%%%%%%%%%%%%%%%%%%%

\begin{abstract}
% Combining traditional and event cameras for robust RGB-E tracking has been a promising research topic in recent years.
The unique complementarity of frame-based and event cameras for high frame rate object tracking has recently inspired some research attempts to develop multi-modal fusion approaches. 
However, these methods directly fuse both modalities and thus ignore the environmental attributes, \eg, motion blur, illumination variance, occlusion, scale variation, etc. Meanwhile, insufficient interaction between search and template features makes distinguishing target objects and backgrounds difficult. As a result, performance degradation is induced especially in challenging conditions.  This paper proposes a novel and effective Transformer-based event-guided tracking framework, called \textbf{eMoE-Tracker}, which achieves new SOTA performance under various conditions. Our key idea is to disentangle the environment into several learnable attributes to dynamically learn the attribute-specific features and strengthen the target information by improving the interaction between the target template and search regions. To achieve the goal, we first propose an environmental Mix-of-Experts (eMoE) module that is built upon the environmental Attributes Disentanglement to learn attribute-specific features and environmental Attributes Assembling to assemble the attribute-specific features by the learnable attribute scores dynamically. The eMoE module is a subtle router that prompt-tunes the transformer backbone more efficiently.  
We then introduce a contrastive relation modeling (CRM) module to emphasize target information by leveraging a contrastive learning strategy between the target template and search regions. Extensive experiments on diverse event-based benchmark datasets showcase the superior performance of our eMoE-Tracker compared to the prior arts. Project page: \url{https://vlislab22.github.io/eMoE-Tracker/}

\vspace{5pt}

\end{abstract}

\vspace{-5pt}
\section{INTRODUCTION}
\IEEEPARstart{V}{isual} object tracking is a critical task with many applications, such as robot scene perception~\cite{ran2021scene} and self-driving~\cite{dai2021tirnet}. It involves tracking the target objects in the sequential video frames based on the initial frame. Many efforts have been made to develop tracking algorithms with standard RGB cameras, however, these methods often fail under challenging conditions,~\eg, low light.

% In recent years,  researchers have attached attention to constructing RGB-E trackers to address the aforementioned challenge. 
Event cameras~\cite{zheng2023deep} are bio-inspired sensors with the merits of high dynamic range and high temporal resolution, which are complementary to conventional RGB cameras. The potential value of the complementarity between the RGB frames and event streams can help improve the robustness of tracking in many challenging visual conditions, \eg, extreme illumination variance and motion blur.
% and cluster, which is crucial for various applications such as surveillance and autonomous driving.

% \IEEEPARstart{T}{he} task of RGB-E tracking has gained much attention in the community recently. The potential value of the complementarity between color frames and event streams can help improve the robustness in many challenging real-world conditions, \eg, extreme illumination variance, motion blur, occlusion, and cluster, which is crucial for various applications such as surveillance and autonomous driving.

\begin{figure}[t!]
\captionsetup{font=small}
\begin{center}
\includegraphics[width=1\linewidth, height=0.4\textwidth]{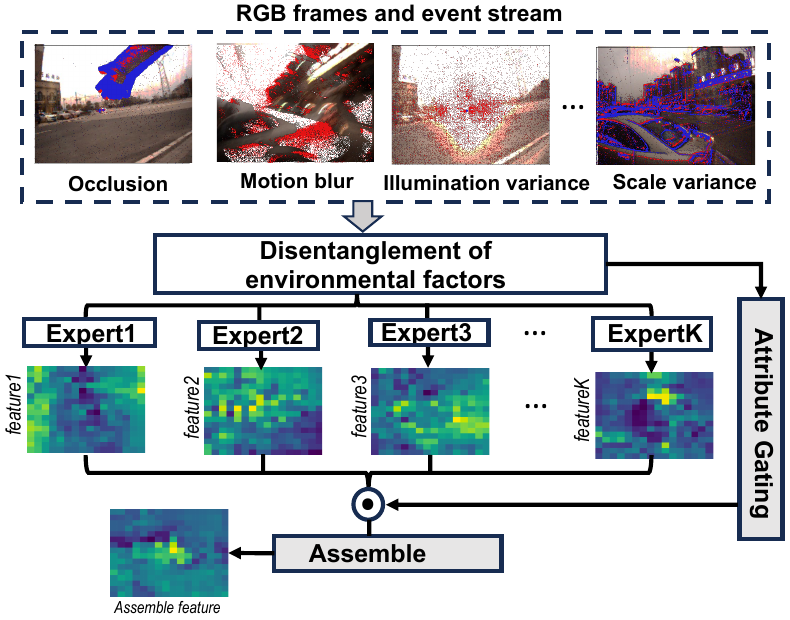}
\end{center}
\vspace{-15pt}
   \caption{An illustration of the core idea of the environmental MoE (eMoE) module. This module acts as a subtle router to prompt-tune the frozen backbone encoder. The number of experts is determined by the attributes we decouple for the environmental conditions, and each expert is responsible for learning the attribute-specific features. All the learned features are assembled and added with the outputs from the backbone encoder at the corresponding layer for robust tracking representation.}
   \vspace{-15pt}
\label{framework_overview}
\end{figure}

This has inspired research endeavors in developing event-guided, \ie, RGB-event (RGB-E) multi-modal tracking approaches~\cite{tang2022revisiting,zhang2021multi,zhang2021object,zhang2023frame,wang2023visevent,zhu2023cross,zhu2023visual}. These works can be divided into two categories based on the network structure: two-stream, \ie, siamese network \cite{zhang2021multi,zhang2021object,zhang2023frame} and one-stream trackers~\cite{tang2022revisiting,zhu2023cross,zhu2023visual}. The former takes two identical branches to process RGB and event modalities separately. To better leverage the complementarity and increase the interaction between them, complex fusion modules are designed, thus leading to model complexity. 
The latter is usually based on the vision transformer (ViT) structure~\cite{dosovitskiy2020image}, where RGB and event tokens are concatenated and fed into the ViT backbone for feature encoding. Although they are free from the complex network structure, they fail to consider the impact of environmental attributes on tracking performance in challenging conditions. 
Meanwhile, inadequate interaction between search and template features makes distinguishing target objects and backgrounds difficult. Consequently, performance degradation is induced especially in challenging conditions.
Intuitively, we raise a novel research question: \textit{how to design a one-stream framework that can distinguish the environmental attributes while enabling feature interaction for robust tracking under diverse visual conditions?}

In this paper, we propose a novel one-stream framework with an environmental Mixture-of-Experts structure (eMoE) along with a contrastive relation modeling (CRM) module to achieve robust tracking in challenging conditions, as shown in Fig.~\ref{framework_overview}. The key insight is to disentangle the environmental attributes through learnable layers
to dynamically learn the attribute-specific features for better tracking representation learning under challenging conditions. 
% interaction between the target objects and background.
Specifically, to disentangle the environmental attributes, the eMoE module ((Sec.~\ref{sec_3.2})) is proposed to achieve two goals: (i) the environmental attributes disentanglement and (ii) the environmental attributes assembling. For the former, 
% the eMoE disentangles the complicated environmental conditions into several learnable attributes to learn the attribute-specific features.
% That is, 
% To explicitly consider the environmental effects on tracking performance,  
the eMoE module disentangles four attributes --illumination variance, motion blur, scale variance, and occlusion --  to learn the attribute-specific features. This has been experimentally shown sufficient to reflect environmental effects on tracking given the advantages of event cameras (See Tab.~\ref{tab:experts}).  
For the latter, each attribute-specific feature is assembled to build a more discriminative representation for tracking~\wrt the attribute scores under the corresponding visual conditions, \eg, motion blur.
For more efficient training, our eMoE module can be inserted into the arbitrary layers to prompt-tune the ViT backbone encoder. For the proposed CRM (Sec.~\ref{sec_4}) module, it aims to better distinguish the target features and background ones by introducing contrastive learning strategy. This subtly improves the interaction between the target template and search region and enhances the target objects. 
By integrating the eMoE and CRM modules, the output features can be more discriminative and less noisy for more robust tracking performance under diverse visual conditions.

To summarize, the contributions of our paper are three-fold: (\textbf{I}) We propose to improve the tracking robustness and precision from the perspective of the environmental attributes. (\textbf{II}) We introduce the environmental Mixture-of-Experts(eMoE) module to disentangle the environment into several learnable attributes for attribute-specific features and assemble them for more discriminative representation in RGB-event tracking tasks. (\textbf{III})  A contrastive relation modeling (CRM) module is designed to further increase the interaction between the search region and target template, thus enhancing the target object information under challenging conditions. 

% \begin{itemize}
%     \item We propose a novel RGB-E tracking model integrating with a visual prompt and a mixture-of-experts structure, which enhances the features under various challenging conditions for more precision tracking results.

%     \item We introduce the dynamic attributes gating module to improve the discriminability and noiselessness under complex real-world conditions.

%     \item A contrastive learning strategy is exploited to further enhance the target object information.
% \end{itemize}

\section{Related Work}
\noindent \textbf{Visual Object Tracking (VOT).}
% It aims at locating and tracking a specific object of interest in a video sequence as it moves over time. Deep learning trackers have dominated the research field of VOT.
% Remarkable progress has been witnessed in the recent study of VOT based on conventional frames. These can be divided into two categories, one is correlation filter trackers\cite{Bertinetto_2016_CVPR,bolme2010visual,henriques2012exploiting,ma2015hierarchical} and the other is deep trackers\cite{bertinetto2016fully, li2018high, nam2016learning, yan2022towards, zhou2022global, zhou2022global1}. 
The mainstream deep trackers can be roughly categorized into two types based on structure: trackers with two-stream networks and with one-stream networks. Siamese-based trackers~\cite{bertinetto2016fully, li2018high, yu2020deformable, chen2020siamese, xu2020siamfc++, Bertinetto_2016_CVPR,zhu2018distractor,wang2019spm} are the archetype two-stream networks, which are designed with two symmetrical branches to learn a similarity function between target template images and search region. On the other hand, trackers with one-stream networks~\cite{cui2022mixformer, chen2022backbone, ye2022joint, lan2023procontext, chen2021transformer} split the target template and search region into a set of tokens, concatenate them, and then feed to a fully-Transformer structure. Among them, MixFormer~\cite{cui2022mixformer} introduces a set of mixed attention modules to extract and integrate the features of the target template and search region simultaneously and to obtain the discriminative target-specific features. OSTrack~\cite{ye2022joint} proposes an early elimination module in the ViT encoder to discriminate the background tokens from the search region. \textit{We utilize the ViT backbone in OSTrack~\cite{ye2022joint} to build a one-stream RGB-E tracker by disentangling the environmental attributes while enabling effective interaction between search and target.}
% for better interaction between the target template and the search region from an early stage.   

\vspace{5pt}
\noindent \textbf{RGB-E Tracking.}
% Thanks to the innate characteristics and complementarity of event cameras, tracking by combining RGB frames and event streams to achieve robust performance in real-world scenarios is promising and reliable. 
Daniel~\etal~\cite{gehrig2018asynchronous, gehrig2020eklt} first tackles the problem of feature tracking using events and frames by developing a maximum likelihood approach based on a generative event model. DashNet~\cite{yang2019dashnet} later achieves an RGB-E tracker by designing the complementary filter and attention module. ESVM~\cite{huang2018event} incorporates event-based guiding methods into the support vector machine to improve tracking accuracy. Recently, Zhang~\etal ~\cite{zhang2021object} introduced self- and cross-domain attention with an adaptive weighting mechanism to fuse frames and events. Tang~\etal~\cite{tang2022revisiting} proposes a one-stream one-stage RGB-E tracking framework to process feature extraction, fusion, matching, and interactive learning simultaneously. ViPT~\cite{zhu2023visual} exploits the modal-relevant prompts to fine-tune the pre-trained backbone model to adapt to multi-modal tracking tasks. However, these methods fail to consider the impact of complex environmental attributes on the tracking performance while only improving the cross-modal fusion. \textit{Differently, our eMoE-Tracker subtly disentangles the environment attributes into several learnable attributes to dynamically learn the attribute-specific features for better interaction and discriminability between the target and background regions}.

%\subsection{Visual Prompt Learning}
\vspace{-5pt}
\section{Method}

% In this section, we propose an environmental prompting network with mixture of experts, namely eMoE-tracker, to effectively and efficiently adapt a pre-trained tracking backbone to RGB-E tracker under real-world challenging situations. The overall architecture comprises an \textbf{V}isual \textbf{P}rompt with \textbf{e}nvironmental \textbf{M}ixture-\textbf{o}f-\textbf{E}xperts module (VP-eMoE) and a \textbf{C}ontrastive \textbf{R}elation \textbf{M}odeling (CRM) module. 

\begin{figure*}[h]
\captionsetup{font=small}
\begin{center}
\includegraphics[width=\linewidth]{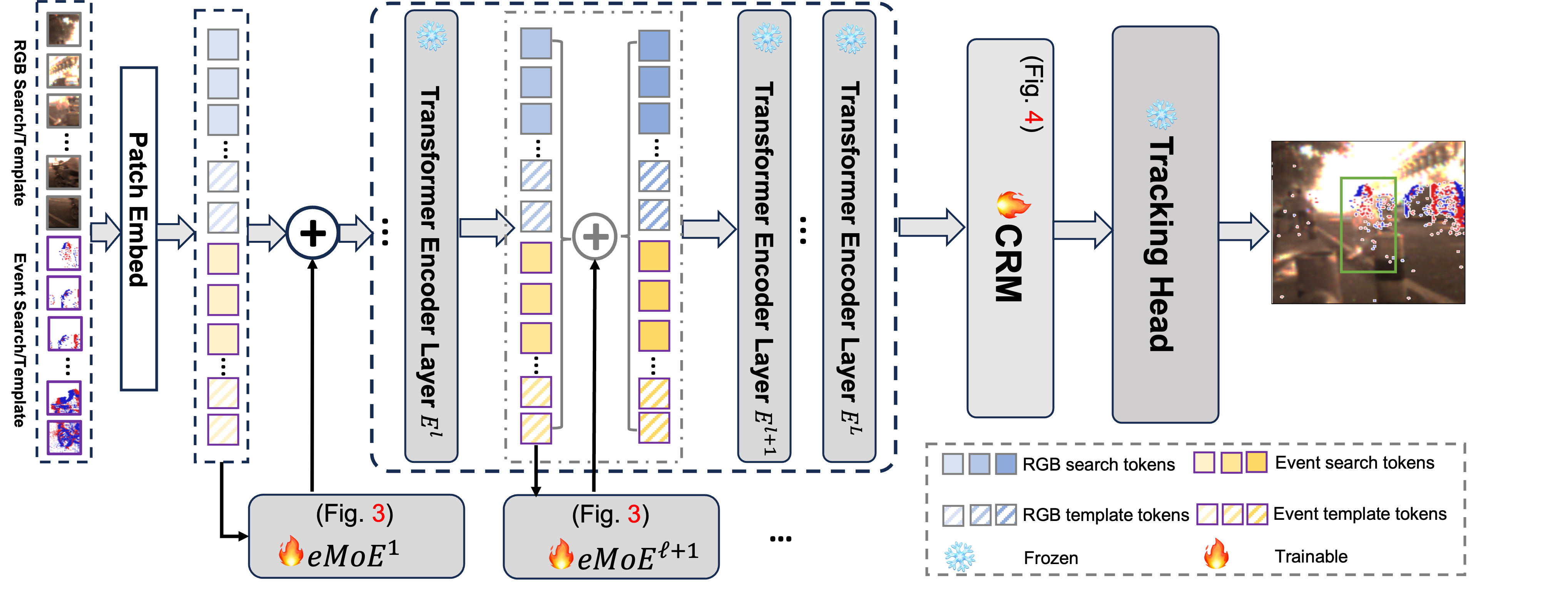}
\end{center}
\vspace{-5pt}
   \caption{Overview of our proposed framework. The input of the whole network is the patch embeddings of RGB frames and stacked event frames. The concatenated two modal patches are fed into the backbone model and eMoE, and eMoE is inserted into the $l$-th layer of ViT to generate feature tokens which are combined with the tokens from the ViT encoder at the corresponding layer. $E^{l}$ is the ViT encoder at layer $l$. The CRM module gets the enhanced tokens to further improve the discriminability of target object.}
\label{framework}
\vspace{-12pt}
\end{figure*}

\subsection{Problem Setting and Overview}

\subsubsection{\textbf{Problem Setting}} Given an initial target bounding box $B_{0}$ in a video, the goal of an RGB-based tracker is to learn a tracking model $T_{RGB}: \{I_{RGB}, B_{0}\} \rightarrow B$ to estimate the bounding box in all subsequent frames $I_{RGB}$. In RGB-E tracking, event streams are introduced and stacked as event frames, extending the input to $(I_{RGB}, I_{E})$, where the subscript $E$ indicates events. For details of representations for event data, we refer readers to \cite{zheng2023deep,gallego2020event}. Therefore, the RGB-E tracking model can be represented as $T_{RGB-E}: \{I_{RGB}, I_{E}, B_{0} \} \rightarrow B$.
% \subsubsection{\textbf{Backbone Model}}
% \label{sec_3}
We choose the transformer encoder and decoder of ~\cite{tang2022revisiting} as our backbone encoder and decoder for better efficiency.
% and early interaction between two modalities. 
The archetypical structure of the backbone encoder and decoder can be represented as $\mathcal{F}_E \circ \mathcal{F}_D$, where $\mathcal{F}_E: \{I_{RGB}, I_{E}, B_{0}\} \rightarrow \mathcal{T}_{RGB-E}$ denotes the backbone encoder and $\mathcal{F}_D: \mathcal{T}_{RGB-E} \rightarrow B$ represents the decoder which produces the estimated bounding box results $B$. The main body of $\mathcal{F}_D$ here is a vanilla vision transformer~\cite{dosovitskiy2020image} containing 12 encoder layers. Each layer contains Multi-head Self-Attention (MSA), LayerNorm (LN), Feed-Forward Network (FFN) and residual connections. Before feeding inputs into the backbone network, RGB and event patches are projected into feature tokens adding with positional embeddings and then concatenated to RGB and event feature tokens $\mathcal{T}_{RGB-E}^{0}=[ \mathcal{T}_{RGB}^{z}, \mathcal{T}_{RGB}^{x},\mathcal{T}_{E}^{z},\mathcal{T}_{E}^{x} ]$ as the inputs of the transformer encoder. Tokens through the $l$-th encoder layer $E^{l}$ can be represented as $\mathcal{T}_{RGB-E}^{l-1}$. The final layer encoder output is denoted as $\mathcal{T}_{RGB-E}^{L}$.  

% The forward propagation function is as follows.

% $$\mathcal{T}_{RGB-E}^{l}=E^{l}(\mathcal{T}_{RGB-E}^{l-1}), \ l=1,2,...,L \eqno{(1)}$$
% $$B= \Phi(\mathcal{T}_{RGB-E}^{L}) \eqno{(2)}$$	

\subsection{The proposed eMoE-Tracker}
\label{sec_3.2}
\subsubsection{\textbf{Overview}} An overview of our eMoE-Tracker is shown in Fig.~\ref{framework}. 
The RGB and event inputs are first projected into a sequence of tokens and fed into backbone encoders and eMoE.
The eMoE module aims to achieve environmental attributes disentanglement and environmental attributes assembling. 
The backbone encoder layers are frozen and the parameters are not updated. 
The eMoE module disentangles the environmental attributes to learn the attribute-specific and assembled features. 
% These features are dynamically added to the tokens from the corresponding layer of the ViT backbone. 
% It dynamically assembles the outputs from eMoE at the corresponding layer.
Outputs from the eMoE module can be dynamically added to the tokens from the corresponding layer of the ViT backbone. 
Overall, the process can be formulated as follows:
$$
     \mathcal{T}^{l} = \mathcal{T}_{RGB-E}^{l} + \mathcal{P}^{l+1}, \ l=1,2,...,L \eqno{(1)}
$$
where $\mathcal{P}^{l+1}$ denotes token features from eMoE at $l+1$ layer.

\vspace{5pt}
\subsubsection{eMoE Module}
The eMoE module aims to achieve: i) environmental attributes disentanglement and ii) environmental attributes assembling. We now describe them.

\begin{figure}[t!]
\captionsetup{font=small}
\begin{center}
\includegraphics[width=1\linewidth]{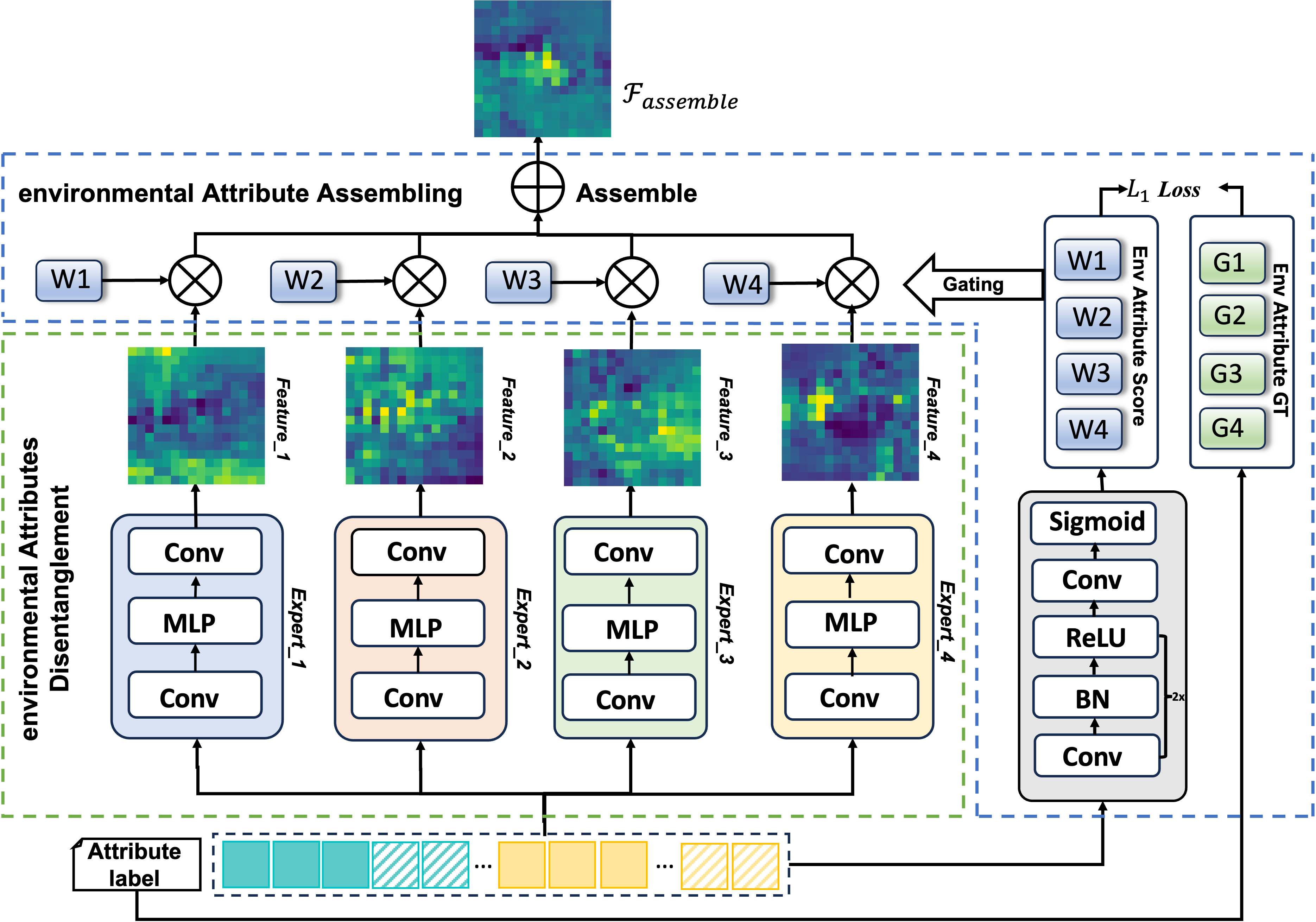}
\end{center}
\vspace{-10pt}
   \caption{An illustration of our eMoE module. Here we take four expert branches as illustrations. The RGB and event tokens are fed into eMoE, which decouples the challenging attributes and generates attribute-specific representations under corresponding challenging conditions. Meanwhile, it is also responsible for dynamically weighing and assembling all the attribute-specific features to form a more discriminative representation for tracking. }

\label{framework_eMoE}
\end{figure}

\vspace{2pt}
\textbf{i. Environmental Attributes Disentanglement.}
%\label{sec_3.2}
As shown in Fig.~\ref{framework_eMoE}, to enable better learning of the attribute-specific features under various conditions, we manually annotate the visible-event datasets with four attribute labels including motion blur, illumination variance, scale variance, and occlusion at video level. Then, a mixture-of-experts network with four identical branches is designed to learn the attribute-specific features for each challenging condition. This allows us to capture more discriminative features and suppress the noises brought by other environmental attributes. All expert networks employ the \textit{CONV-MLP-CONV} structure but with different parameters. Considering the $l$-th ViT layer, we assume that there are $K$ experts $\{f_{expert}^{l, i}(\mathcal{T}^{l}):\mathbb{R}^{N\times D }\rightarrow \mathbb{R}^{N\times D}, i \in [1, K]\}$ to learn the attribute-specific features under the corresponding environmental condition, where $l$ denotes the layer of ViT backbone and $i$ represents the index of experts. Through each expert, we generate a series of attribute-specific features $\{\mathcal{H}_{i}^{l} \in \mathbb{R}^{N\times D}, i \in [1, K] \}$ by expert $f_{expert}^{l, i} (\mathcal{T}^{l}) 
$.

\vspace{5pt}
\textbf{ii. Environmental Attribute Assembling.}
% \label{sec_3.3}
After obtaining the attribute-specific features under different decoupled environmental conditions, we should consider the different contributions of these features with the supervision of the ground truth attribute labels $G = [G_{1}, G_{2}, ... , G_{K}]$. To achieve the goal, the assembling network employs the \textit{CONV-BatchNorm-ReLU-CONV-Sigmoid} structure with two loops and is fed by all the RGB and event patch tokens to generate $K$ attribute scores for attribute-specific features, where $K$ denotes the number of experts. The learnable score indicates the ratio of different challenging types in the corresponding scenario, therefore the attribute-specific feature with a larger score should have a higher contribution to the assembling features to achieve the robust representation under various challenging conditions. Moreover, it can suppress the noise from other environmental attributes. Specifically, at $l$-th layer of the backbone, the attribute scores $W^{l, t}$ are generated from the assembling network $\{f_{a}^{l, t}(\mathcal{T}^{l}):\mathbb{R}^{N\times D} \rightarrow \mathbb{R}^{K} \}$, where $t$ represent the index of experts. The assembling feature $\mathcal{F}_{assemble}^{l}$ at layer $l$ can be formally calculated by $
\sum_{t=1}^{K}W^{l,t}\mathcal{H}_{t}^{l}. 
$ 

\subsubsection{Contrastive Relation Modeling}
\label{sec_4}
\begin{figure}[t!]
\captionsetup{font=small}
\begin{center}
\includegraphics[width=1\linewidth]{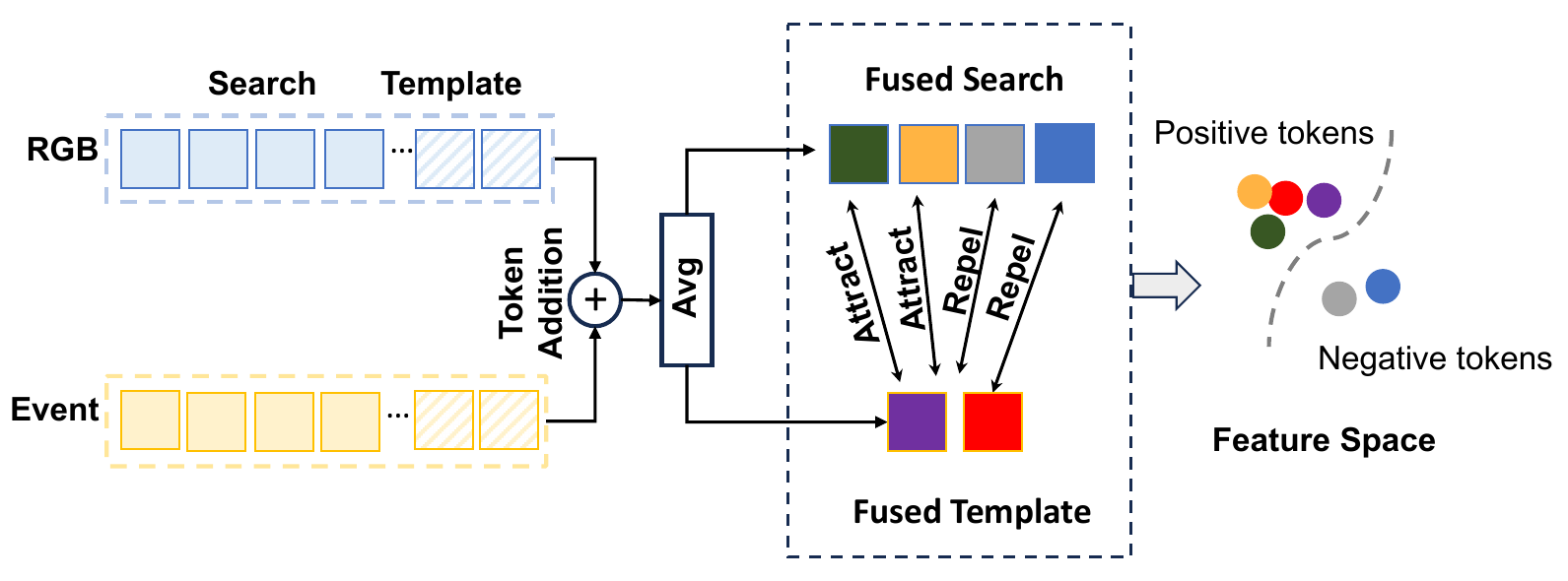}
\end{center}
\vspace{-15pt}
   \caption{An illustration of our CRM module. The RGB and event tokens are first fused into fused search region feature tokens and target template feature tokens. We exploit the contrastive learning strategy to pull the target information in search tokens near template feature tokens while push background information away from template feature tokens. The final goal is to make the tracking features more discriminative and unambiguous.}
   \vspace{-15pt}
\label{framework_CRM}
\end{figure}

Apart from increasing the distinguishable ability of representations in different environmental conditions, we want to enhance the target information by introducing more interaction between target template and search region. Based on the fact that the features from the target template primarily contains target information, while the search regions include both the target and background information, we propose a CRM module by leveraging the contrastive learning strategy. As shown in Fig.~\ref{framework_CRM}, we first fuse two-modal patch tokens into fused target template tokens and search region tokens to better build the relationship. After fusion, we create positive pairs between features from target template and target object contents of search regions, and negative pairs between features from target template and background contents of search regions. To better strengthen the target object information, contrastive learning mechanism is exploited to pull the positive pairs closer and push negative pairs away thereby allowing the target object and background more distinguishable. The proposed CRM module effectively helps generate more unambiguous representation and achieve high performance in various challenging conditions.
% Apart from the eMoE module helping increase the discriminative ability of features under complicated scenes, we propose a CRM module to increase the interaction between search region and target template and enhance target information. To achieve it, let us assume that the features from the target template contain mainly target information. In contrast, the features from the search region include both the target and background information. We first fuse corresponding patch tokens into fused target template tokens and search region tokens to better build the relation on two modal data. After fusion, we create positive pairs between features of the target template and target features of the search region, and negative pairs between features of the target template and background features of the search region as shown in Fig.~\ref{framework_CRM}. This helps us pull the target template tokens near the target-related tokens while pushing background-related tokens away from the search region, thus improving the discriminability. 

\vspace{-5pt}
\subsection{Optimization}
The main body of our RGB-E tracker $F_{RGB-E}$ is initialized by the transformer-based tracking backbone~\cite{ye2022joint}. All the parameters $\theta$ we should update is only existing in eMoE and CRM. The optimization process can be formulated as
$$
\theta = arg min \frac{1}{|\mathcal{D}|}\sum\mathcal{L}(CRM(\mathcal{F}_D(\mathcal{T}_{RGB-E}^{L})), B_{gt}), \ \eqno{(2)}
$$
where $|\mathcal{D}|$ denotes the RGB-event data.

The overall objective function of our model includes tracking loss $L_{tracking}$, contrastive loss $L_{NCE}$ and attribute loss $L_{attr}$. The tracking loss is the same as the transformer-based tracking backbone~\cite{ye2022joint} as follows,
$$
   L_{tracking} = L_{cls}+ \lambda_{iou}L_{iou} + \lambda_{L_{1}}L_{1} \ \eqno{(3)} 
$$
where $L_{cls}$ is the focal loss~\cite{law2018cornernet} for classification, IoU loss~\cite{rezatofighi2019generalized} $L_{iou}$ and $L_{1}$ are exploited for bounding box regression, $\lambda_{iou}$ and $\lambda_{L_{1}}$ are regularization parameters. For more details, please refer to~\cite{ye2022joint}.
Additionally, we take the InfoNCE loss~\cite{jaiswal2020survey} as a contrastive learning loss for the CRM module. Given the fused target template tokens $\mathcal{T}_{fused}^{z} = [t_{fused}^{z, 1}, t_{fused}^{z, 2}, ..., t_{fused}^{z, N_{z}}]$ from the ViT encoder, we compute the similarity $S=[s^{1},s^{2},...,s^{N_{x}}]$ between $\mathcal{T}_{fused}^{z}$ and the fused search region tokens $\mathcal{T}_{fused}^{x}=[t_{fused}^{x, 1}, t_{fused}^{x, 2}, ..., t_{fused}^{x, N_{x}}]$.
where $\tau$ is the temperature parameter and $s^{i} = sim(\mathcal{T}_{fused}^{z}, t_{fused}^{x, i})/\tau 
$.
Based on that, the search region tokens which contain the information inside the ground-truth bounding box can be selected as positive pairs including $N_{pos}$, where the similarity score is defined as $[s_{p}^{k}]_{k=1}^{N_{pos}}$. The left are negative pairs including $N_{neg}$, where the similarity score is set as $[s_{n}^{k}]_{k=1}^{N_{neg}}$. 
The contrastive learning loss can be formulated as:
$$
\mathcal{L}_{NCE}=-log(\frac{\sum_{k=1}^{N_{pos}}e^{s_{p}^{k}}}{\sum_{k=1}^{N_{pos}}e^{s_{p}^{k}}+\sum_{k=1}^{N_{neg}}e^{s_{n}^{k}}}) \ \eqno{(4)}
$$
% $$
%   L_{NCE} = -log\frac{exp(q \cdot k_{+}/\tau)}{\sum_{i=0}^{k}exp(q\cdot k_{i}/\tau)}
% $$
For the attribute loss $L_{attr}$, we utilize $L_{1}$ term to measure the distance between the estimated attribute scores and the ground truth labels. It can be formulated as:
$$
  L_{attr} = \sum_{t,l}||W^{l,t} - G^{t}||_{l_{1}}  \ \eqno{(5)}
$$

The total objective can be formulated as follows:
$$
  \mathcal{L} = L_{tracking} + \alpha L_{NCE} + \beta L_{attr} \ \eqno{(6)}
$$

\begin{table*}[th]
\centering
\caption{Experimental results on VisEvent dataset. The best results are shown in \textbf{bold}. }
\vspace{-5pt}
\resizebox{\textwidth}{!}{
\begin{tabular}{c|llllllllllllll}
\toprule
Tracker& \multicolumn{1}{c}{Ocean~\cite{zhang2020ocean}} & \multicolumn{1}{c}{SiamCAR~\cite{guo2020siamcar}} & \multicolumn{1}{c}{SiamRPN++~\cite{li2019siamrpn++}} & \multicolumn{1}{c}{ATOM~\cite{danelljan2019atom}} & \multicolumn{1}{c}{PrDiMP~\cite{danelljan2020probabilistic}} & \multicolumn{1}{c}{LTMU~\cite{dai2020high}} &\multicolumn{1}{c}{SiamAPN++~\cite{cao2021siamapn++}}& {EFTrack~\cite{zhang2023eftrack}} & \multicolumn{1}{c}{FENet~\cite{zhang2021object}}  & \multicolumn{1}{c}{AFNet~\cite{zhang2023frame}} & \multicolumn{1}{c}{OSTrack~\cite{ye2022joint}}  
& \multicolumn{1}{c}{CEUTrack~\cite{tang2022revisiting}}  & \multicolumn{1}{c}{ViPT~\cite{zhu2023visual}}  &\multicolumn{1}{c}{\textbf{eMoE-Tracker(Ours)}}    \\
  % &~\cite{siamrpn}      &~\cite{siamban} & ~\cite{siamfc++} & ~\cite{kys}  &~\cite{clnet}  &~\cite{wang2023visevent}   &~\cite{atom}   &~\cite{dimp}   &~\cite{prdimp}   &~\cite{wang2023visevent}   &       \\
\midrule
\hline
SR & \multicolumn{1}{c}{23.26} & \multicolumn{1}{c}{34.49} & \multicolumn{1}{c}{33.66} & \multicolumn{1}{c}{31.34} & \multicolumn{1}{c}{37.39} & \multicolumn{1}{c}{37.05} & \multicolumn{1}{c}{42.7}&\multicolumn{1}{c}{43.6} & \multicolumn{1}{c}{44.2} & \multicolumn{1}{c}{44.5}   & \multicolumn{1}{c}{53.4}   & \multicolumn{1}{c}{55.58}  & \multicolumn{1}{c}{59.2} & \multicolumn{1}{c}{\textbf{61.3}} \\

PR & \multicolumn{1}{c}{52.02} & \multicolumn{1}{c}{58.86} & \multicolumn{1}{c}{60.58} & \multicolumn{1}{c}{60.45} & \multicolumn{1}{c}{64.47} & \multicolumn{1}{c}{66.76} & \multicolumn{1}{c}{56.2} &\multicolumn{1}{c}{57.3} &\multicolumn{1}{c}{58.9} & \multicolumn{1}{c}{59.3}   & \multicolumn{1}{c}{69.5}   & \multicolumn{1}{c}{69.06 }  & \multicolumn{1}{c}{ 75.8}  & \multicolumn{1}{c}{\textbf{76.4 }}  \\

NPR & \multicolumn{1}{c}{54.21} & \multicolumn{1}{c}{62.99} & \multicolumn{1}{c}{64.72} & \multicolumn{1}{c}{63.41} & \multicolumn{1}{c}{67.02} & \multicolumn{1}{c}{69.78} & \multicolumn{1}{c}{56.8} &\multicolumn{1}{c}{58.0} &\multicolumn{1}{c}{61.2} & \multicolumn{1}{c}{62.5}   & \multicolumn{1}{c}{72.6}   & \multicolumn{1}{c}{73.0}  & \multicolumn{1}{c}{73.2}  & \multicolumn{1}{c}{\textbf{79.6}}  \\

\bottomrule
\end{tabular}
} 
\label{viseventtable}
\end{table*}

\vspace{-15pt}

\begin{table*}[th]
\centering
\caption{Experimental results on COESOT dataset. The best results are shown in \textbf{bold}. }
\vspace{-5pt} 
\resizebox{\textwidth}{!}{
\begin{tabular}{c|llllllllllllll}
\toprule
Tracker& \multicolumn{1}{c}{MixFormer1k~\cite{cui2022mixformer}}  & \multicolumn{1}{c}{STARK-S50~\cite{yan2021learning}}  & \multicolumn{1}{c}{PrDiMP50~\cite{danelljan2020probabilistic}}  & \multicolumn{1}{c}{PrDiMP18~\cite{danelljan2020probabilistic}}  & \multicolumn{1}{c}{ATOM~\cite{danelljan2019atom}}  &  \multicolumn{1}{c}{SiamRPN~\cite{li2018high}}  & \multicolumn{1}{c}{AiATrack~\cite{gao2022aiatrack}}  & \multicolumn{1}{c}{TrSiam~\cite{wang2021transformer}} & 
\multicolumn{1}{c} {SiamAPN++~\cite{cao2021siamapn++}}& {EFTrack~\cite{zhang2023eftrack}}&\multicolumn{1}{c}{OSTrack~\cite{ye2022joint}}  
& \multicolumn{1}{c}{CEUTrack~\cite{tang2022revisiting}}  & \multicolumn{1}{c}{ViPT~\cite{zhu2023visual}} &
\multicolumn{1}{c}{\textbf{eMoE-Tracker(Ours)}}    \\
  % &~\cite{siamrpn}      &~\cite{siamban} & ~\cite{siamfc++} & ~\cite{kys}  &~\cite{clnet}  &~\cite{wang2023visevent}   &~\cite{atom}   &~\cite{dimp}   &~\cite{prdimp}   &~\cite{wang2023visevent}   &       \\
\midrule
\hline
SR  & \multicolumn{1}{c}{56.0} & \multicolumn{1}{c}{55.7} & \multicolumn{1}{c}{57.9} & \multicolumn{1}{c}{56.7} & \multicolumn{1}{c}{55.0} & \multicolumn{1}{c}{53.5} & \multicolumn{1}{c}{59.0} & \multicolumn{1}{c}{59.7}   &\multicolumn{1}{c}{57.8}  & \multicolumn{1}{c}{58.4} & \multicolumn{1}{c}{59.0}   & \multicolumn{1}{c}{62.7}  & \multicolumn{1}{c}{65.3}&\multicolumn{1}{c}{\textbf{67.1}} \\

PR  & \multicolumn{1}{c}{62.8} &\multicolumn{1}{c}{62.6} & \multicolumn{1}{c}{65.0} & \multicolumn{1}{c}{62.9} & \multicolumn{1}{c}{63.6} & \multicolumn{1}{c}{61.1} & \multicolumn{1}{c}{67.4} & \multicolumn{1}{c}{66.3} & 
\multicolumn{1}{c}{60.1}  & \multicolumn{1}{c}{65.2}  & \multicolumn{1}{c}{66.6}   & \multicolumn{1}{c}{70.9 }  & \multicolumn{1}{c}{ 73.7}  &\multicolumn{1}{c}{\textbf{79.9}}  \\

NPR  & \multicolumn{1}{c}{61.7} &\multicolumn{1}{c}{61.6} & \multicolumn{1}{c}{ 64.0} & \multicolumn{1}{c}{ 62.6} & \multicolumn{1}{c}{ 63.0} & \multicolumn{1}{c}{76.4 }  & 
\multicolumn{1}{c}{62.7} & \multicolumn{1}{c}{ 62.8} & \multicolumn{1}{c}{60.6} & \multicolumn{1}{c}{61.1}   & \multicolumn{1}{c}{62.8}   & \multicolumn{1}{c}{73.9 }  & \multicolumn{1}{c}{65.9} &
\multicolumn{1}{c}{\textbf{ 82.3}}  \\
\bottomrule
\end{tabular}
} 
\label{coesottable}
\end{table*}

\section{Experiment}

% \begin{figure*}[h]
% \captionsetup{font=small}
% \begin{center}
% \includegraphics[width=\linewidth]{Images/Fig_attention_map.pdf}
% \end{center}
% \vspace{-18pt}
%    \caption{Visualization of attention maps from the layers of backbone ViT, ranging from layer 7 to layer 5. The first and third rows are the attention maps from the foundation model OSTrack, and the second and fourth rows are from our method VPAT.}
% \label{visualization}
% \vspace{-12pt}
% \end{figure*}

\begin{figure}[t!]
\captionsetup{font=small}
\begin{center}
\includegraphics[width=1\linewidth]{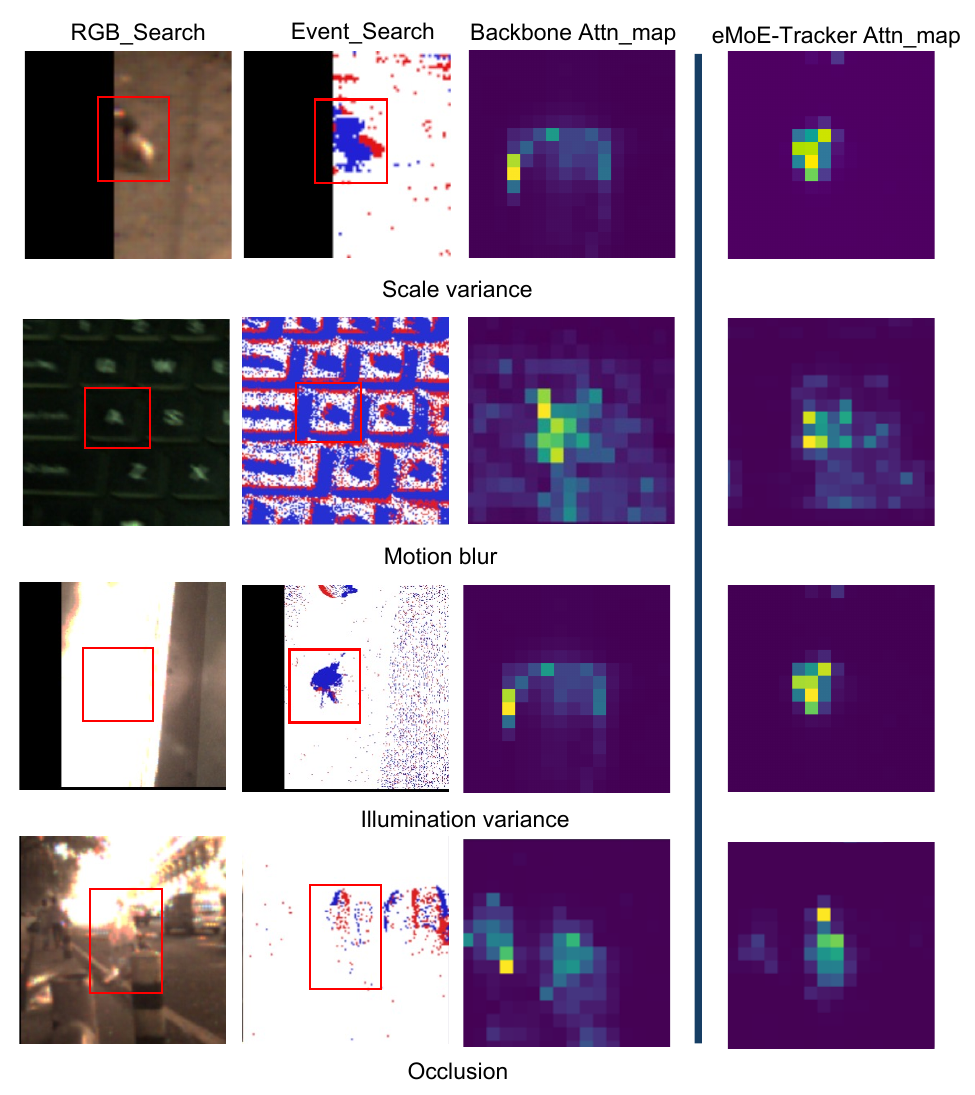}
\end{center}
\vspace{-15pt}
   \caption{Visualization of attention maps from the backbone network compared with our eMoE-Tracker. Four challenging conditions including illumination variance, motion blur, occlusion and scale variance are selected to reflect the effectiveness of environmental attributes disentanglement and feature assembling.}
   \vspace{-15pt}
\label{visualization}
\end{figure}

\subsection{Experimental Settings}
 Our method is trained end-to-end on 1 NVIDIA A800 GPU with PyTorch implementation. During training, our method utilizes a global batch size of 64 and takes 60 epochs, and each epoch processes $6\times 10^{4}$ sample pairs. We employ the AdamW~\cite{loshchilov2017decoupled} optimizer with a weight decay of $10^{-4}$ and set the initial learning rate as $2\times 10^{-4}$ but decreasing after 32 epochs by the factor of 10. 
 % The initial parameters of ViT are loaded from the pre-trained backbone model~\cite{ye2022joint} in RGB-based tracking and the trainable prompt branch is initialized following a truncated normal distribution. 

We use two datasets to demonstrate the effectiveness of our method: VisEvent~\cite{wang2023visevent} and COESOT~\cite{tang2022revisiting}. We compare with some RGB-E trackers, including ViPT~\cite{zhu2023visual}, CEUTrack~\cite{tang2022revisiting}, FENet~\cite{zhang2021object}, AFNet~\cite{zhang2023frame} and many other RGB-based trackers with two-modal input, \eg, SiamRPN++~\cite{li2019siamrpn++}, ATOM~\cite{danelljan2019atom}, STARK~\cite{yan2021learning}, MixFormer~\cite{cui2022mixformer}, etc.
We adopt three metrics to evaluate trackers' performance, including {precision rate (PR), success rate (SR), and normalized precision rate (NPR)}.
% VisEvent is one of the visible-event benchmark datasets which contains 500 video sequences for training and 320 video sequences for testing. Compared with VisEvent~\cite{wang2023visevent}, COESOT~\cite{tang2022revisiting} is the current largest RGB-E dataset which contains 1354 color-event videos with 478,721 RGB frames. The videos are captured from real-world scenarios including indoor and outdoor conditions. There are 827 video sequences for training and 527 videos for testing.

% \textbf{Evaluation Metric.}  Specifically, the success rate (SR) is typically calculated as the percentage of frames in which the tracked object is successfully located and tracked within a specified region of interest. The precision rate (PR) measures the percentage of frames in which the tracked object is correctly located within a certain distance from its ground truth position. NPR denotes normalized precision rate.

\begin{figure}[t!]
\captionsetup{font=small}
\begin{center}
{\includegraphics[width=\linewidth]{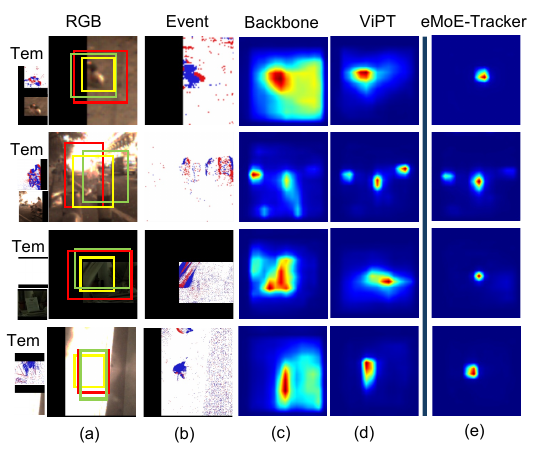}}
\end{center}
\vspace{-15pt}
   \caption{Visualization on the score head maps from backbone model, ViPT~\cite{zhu2023visual} and our model eMoE-Tracker. The \textcolor{red}{red}, \textcolor{green}{green} and \textcolor{yellow}{yellow} boxes denote the bounding box of the backbone network, ViPT~\cite{zhu2023visual} and eMoE-Tracker, respectively. (a)RGB search frame. (b)Stacked event search frame. (c)Score maps from the backbone network. (d) Score maps from ViPT. (e) Score maps from eMoE-Tracker.}
   \vspace{-15pt}
\label{scorehead}
\end{figure}

\vspace{5pt}
\begin{figure}[t!]
\captionsetup{font=small}
\begin{center}
\includegraphics[width=1\linewidth]{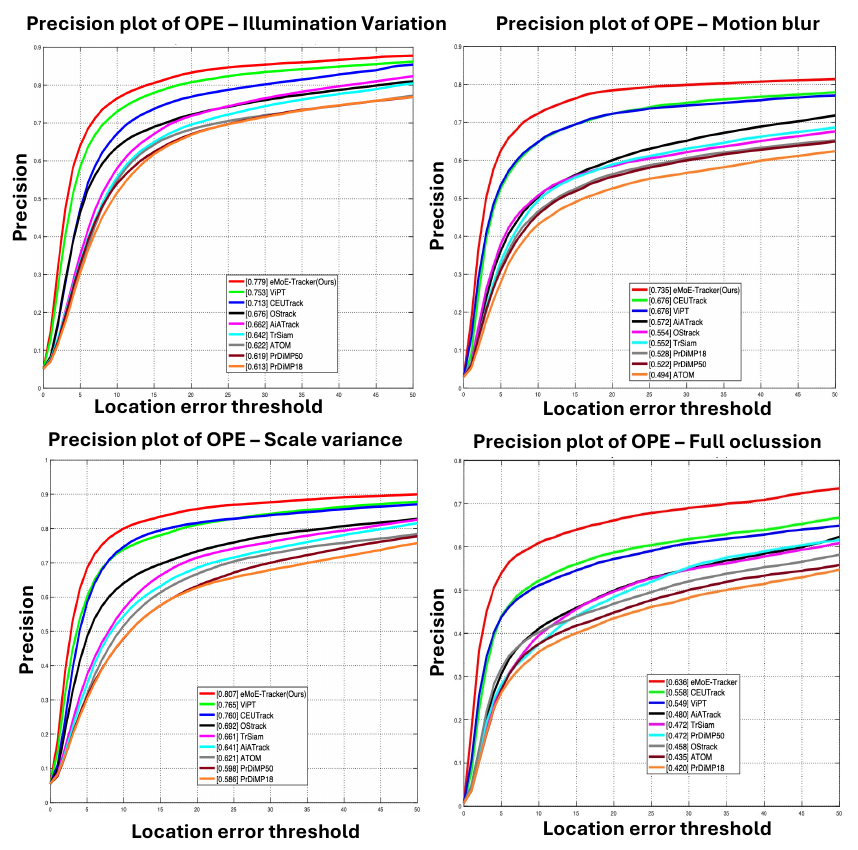}
\end{center}
\vspace{-15pt}
   \caption{Overall performance on COESOT under four challenging attributes, including illumination variation, motion blur, scale variance and full occlusion.}
\label{Attributes}
\vspace{-12pt}
\end{figure}

\vspace{-15pt}
\subsection{Comparison}

\textbf{Evaluation on VisEvent.}
We evaluate our method on the VisEvent dataset compared to SOTA trackers with two modal inputs. Note that we employ the stacked event frames while not the raw event streams as input in our model. The quantitative results are illustrated in TABLE~\ref{viseventtable}. Our method is superior to other SOTA trackers, which achieve 61.3\%, 76.4\%, and 79.6\% on the metrics of SR, PR, and NPR, respectively. Surprisingly, our method surpasses the backbone network by 6.2\% and 6.9\% on SR and PR and exceeds existing RGB-E SOTA ViPT by 0.4\% and 0.6\% respectively on SR and PR, which demonstrates the effectiveness of our method. 

\textbf{Evaluation on COESOT.}
COESOT is the largest real-world visible-event benchmark dataset. We compare with 12 RGB-E trackers to evaluate the effectiveness of our method. We report our results in TABLE~\ref{coesottable}. As observed, our method achieves the best performance among all the trackers, with the figure of 67.1\%, 79.9\%, and 82.3\% on SR, PR, and NPR. Additionally, eMoE-Tracker shows a gain of 1.8\% on SR and 6.2\% on PR respectively, and also outperforms the backbone network by a large margin. It demonstrates that our algorithm achieves the SOTA performance on the COESOT dataset.

\vspace{5pt}
\textbf{Visualization.} Qualitative results are provided in Fig.~\ref{visualization} and Fig.~\ref{scorehead}. Specifically, Fig.~\ref{visualization} shows the attention maps from the backbone network and eMoE-Tracker, where our model can generate a more discriminative response under some complex scenarios, \eg, scale variance.  In Fig.~\ref{scorehead}, a more precise location of the target can be provided by eMoE-Tracker compared with the backbone network and ViPT~\cite{zhu2023visual}.

\vspace{-10pt}
\begin{table}[ht]
	\centering
    \caption{Ablation studies on the effectiveness of our proposed modules: eMoE and CRM, and tracking header frozen or not. 
	}
	\fontsize{4}{6}\selectfont  
	\setlength{\tabcolsep}{1.6mm}{
		\resizebox{\linewidth}{!}{%
			\begin{tabular}{c|ccc|cc|cc}
				\hline
                
				\multirow{2}{*}{Model} &
				
                \multirow{2}{*}{eMoE} &
				\multirow{2}{*}{CRM} &
                   \multirow{2}{*}{Header Unfrozen} &
               
				\multicolumn{2}{c}{VisEvent} &
				\multicolumn{2}{c}{COESOT}  \\
				\cline{5-6}
				\cline{7-8}
				&&  &&SR & PR & SR & PR\\
				\hline
				Backbone & &&& 53.4 & 69.5 & 59.0 & 66.6  \\
				
                \ding{172} & \yes&&&59.2&75.8 &65.8 &75.0 \\
				\ding{173} &\yes&\yes &&\textbf{61.3}  &\textbf{76.4}& \textbf{67.1} & \textbf{79.9}  \\       
                \ding{174} & \yes&\yes&\yes&59.0&74.2&63.8&74.8 \\
				\hline
	\end{tabular}}}
	\vspace{1.5mm}
	
	\vspace{-5mm}
	\label{tab:ablation_study}
\end{table}

\vspace{-10pt}
\begin{table}[ht]
	\centering
    \caption{Ablation studies on the number of experts.
	}
    \scalebox{1.0}{
        \fontsize{3}{5}\selectfont  
	\setlength{\tabcolsep}{1.6mm}{
		\resizebox{\linewidth}{!}{%
			\begin{tabular}{c|cc|cc}
				\hline
                
				\multirow{2}{*}{The number of experts} &
               
				\multicolumn{2}{c}{VisEvent} &
				\multicolumn{2}{c}{COESOT}  \\
				\cline{2-3}
				\cline{4-5}
				&SR & PR & SR & PR\\
				\hline
                1 & 54.2 &70.8& 60.6 &67.1   \\
				2 & 58.6 &71.6& 62.1 &70.9   \\
				3&59.0&73.5 & 63.4&72.6 \\
                4&\textbf{61.3}&\textbf{76.4}&\textbf{67.1}&\textbf{79.9} \\
                
				\hline
	\end{tabular}}}
    }
	
	\vspace{-5mm}
	\label{tab:experts}
\end{table}

\vspace{5pt}
\subsection{Ablation Studies}
\textbf{Effectiveness of eMoE and CRM.} To validate the effectiveness of our proposed modules, we perform the ablation studies on the VisEvent and COESOT datasets. We implement four comparison experiments inside the network. They are: 1) backbone network 2) backbone network with eMoE; 3) backbone network with eMoE and CRM. The ablation studies can be found in TABLE~\ref{tab:ablation_study}. As observed, the best performance happens when we combine all the proposed modules into the backbone network. For the VisEvent dataset, with the incorporation of eMoE and CRM, our method outperforms the backbone model by 7.9\% and 6.9\% on metrics SR and PR, respectively. Additionally, adding the CRM module to model \ding{172}, the SR and PR gain improvements of 2.1\% and 0.6\%, respectively. The results showcase the effectiveness of the proposed modules.
 
% What's more, comparing experiment \ding{174} and \ding{175}, we can find incorporating the CRM module can further increase the precision of tracking, which shows the contrastive learning strategy can increase the interaction between search region and target template and enhance the target information.

\textbf{Analysis on environmental attributes.}
In the dataset COESOT, there are 17 challenging environmental attributes annotated to help analyze the performance under different challenging conditions. Here we illustrate the overall performance on the COESOT dataset for the disentangled four challenging attributes: illumination variation, motion blur, scale variance, and full occlusion. In Fig.~\ref{Attributes}, we can find our proposed model eMoE-Tracker outperforms the backbone model and ViPT~\cite{zhu2023visual}. Specifically, it achieves 63.6\% in occlusion, 77.9\% in illumination variance, 73.5\% in motion blur, and 80.7\% in scale variance on PR, respectively. The results demonstrate that our algorithm is effective in improving tracking precision and robustness under various challenging scenarios.

\textbf{Inserted layers of eMoE.}
We achieve RBG-E tracking under various challenging conditions by injecting eMoE blocks into different layers of the backbone model. It is intuitive to investigate the effect on the number of inserted blocks. Here we set different insert intervals for blocks to insert and the intervals are 1,2 4, 6, and 12. Therefore, the first means that all the layers are fully inserted and the last one only inserts the blocks in the last high-level layer. As shown in Fig.~\ref{tab:Inserted_blocks}, the best performance can be obtained when the tracking backbone network is fully inserted.

\vspace{-12pt}
\begin{table}[ht]
	\centering
    \caption{Ablation studies on inserted intervals into the backbone encoders.
	}
   \scalebox{1.0}{
   \fontsize{3}{5}\selectfont  
	\setlength{\tabcolsep}{1.6mm}{
		\resizebox{\linewidth}{!}{%
			\begin{tabular}{c|cc|cc}
				\hline
                
				\multirow{2}{*}{Inserted intervals} &
               
				\multicolumn{2}{c}{VisEvent} &
				\multicolumn{2}{c}{COESOT}  \\
				\cline{2-3}
				\cline{4-5}
				&SR & PR & SR & PR\\
				\hline
				1 &\textbf{61.3} &\textbf{76.4} &\textbf{67.1} &\textbf{79.9}   \\
				2& 60.0& 75.8 &66.3  &76.1  \\
                4&58.1 &72.6 &63.4 &72.5  \\
				6&55.8 &72.0 &61.1  &70.9   \\
                12&54.7 &70.3&60.3 &68.0 \\
				\hline
	\end{tabular}}}
   }
	
	\vspace{1.5mm}
	\vspace{-5mm}
	\label{tab:Inserted_blocks}
\end{table}

\vspace{15pt}
\textbf{Analysis on the number of experts.}
Due to the complicated environmental attributes, it is worthwhile to consider the impact of the number of experts on tracking performance. Intuitively, more experts are more powerful at addressing complex environments and decomposing them into environmental attributes for easier tracking. 
However, on the one hand, too many experts increase the burden on the model and might result in overfitting, on the other hand, the number of experts is unfeasible to extend over four due to the limitation of manual annotations. Therefore, we conduct ablation studies on the number of experts, and the results can be found in TABLE~\ref{tab:experts}. 

\vspace{5pt}
\textbf{Analysis on the model complexity.} As we mentioned previously, trackers with the two-stream structure, \eg siamese-based trackers, suffer from model complexity due to the high demand for multi-modal fusion. To evaluate the superiority of one stream tracker on network complexity, we calculate the number of trainable parameters on some two stream trackers, like AFNet~\cite{zhang2023frame} and FENet~\cite{zhang2021object}, and our proposed eMoE-Tracker for the comparison. Results are reported in TABLE~\ref{parameter}.

\vspace{-5pt}
\begin{table}[t!]
\centering
\caption{Trainable parameters comparison among two stream trackers, one stream trackers, and our eMoE-Tracker. OS and TS denote one stream and two streams, respectively.}
\fontsize{7}{9}\selectfont  
\resizebox{\linewidth}{!}{
\begin{tabular}{c|llllll}
\toprule
Tracker&AFNet~\cite{zhang2023frame}&FENet~\cite{zhang2021object}&VisEvent~\cite{wang2023visevent}&ViPT~\cite{zhu2023visual}&CEUTrack~\cite{tang2022revisiting}&eMoE-Tracker \\
\hline
Structure-type&TS&TS&TS&OS&OS&OS \\
Trainable Parameters(MB)&25.16&41.87&27.53&0.84&93.7&8.42\\
\bottomrule
\end{tabular}}
\vspace{-15pt}
\label{parameter}
\end{table}

% \vspace{-3pt}
\section{CONCLUSIONS}

In this work, we proposed eMoE-Tracker, a one-stream transformer-based tracking model by introducing mixture-of-experts structure and contrastive learning scheme to RGB-E tracker under various challenging conditions. 
% The core idea is to exploit multiple experts to decouple the challenging environment into learnable attributes for attribute-specific features and assemble them for more discriminative representations for tracking in challenging conditions. Moreover, we introduced a contrastive relation modeling module to further enhance the target information representation, thus improving the tracking performance. 
Extensive experiments on benchmark visible-event datasets VisEvent and COESOT demonstrate the robustness and effectiveness of eMoE-Tracker for RGB-E tracking under challenging conditions like motion blur, illumination variance and etc. We can gain insights from the results that the tracking performance degradation in challenging conditions can be alleviated by explicitly considering tracking tasks from an environmental attributes perspective.

\noindent \textbf{Limitations.} Despite the superior performance for RGB-E tracking, the limitation of our model is highly dependent on the manual video-level attribute annotations for the environmental attributes, thus putting restrictions on the generalization of the model. Also, video-level annotations might not provide absolute precise labels for every video sequence in some conditions. 
% Additionally, the real-world scenarios are complicated and hard to fully consider all the environmental attributes manually. 
In the future, we expect to learn an agent to obtain the environmental attributes in a learnable manner for multi-modal tracking tasks.

%%%%%%%%%%%%%%%%%%%%%%%%%%%%%%%%%%%%%%%%%%%%%%%%%%%%%%%%%%%%%%%%%%%%%%%%%%%%%%%%

% \begin{thebibliography}{99}

\newpage
\bibliographystyle{ieeetr}
\bibliography{root}

 \clearpage
\begin{center}
    {\LARGE \bf  Appendix}
\end{center}

\vspace{5pt}

\begin{abstract}
% Combining traditional and event cameras for robust RGB-E tracking has been a promising research topic in recent years.
Due to the limited space in the main paper, we provide additional material for the proposed method and experimental results. Sec.~\ref{sec:dataset and implementation} introduce the datasets. Then, Sec.~\ref{sec.experiments} illustrates more details about the implementation and experiments. Afterward, we report more visual results and performance evaluation under different attributes in Sec.~\ref{sec:additional results}. In the end, Sec.~\ref{sec:conclusion} makes summary for the complementary material.
\end{abstract}

%%%%%%%%%%%%%%%%%%%%%%%%%%%%%%%%%%%%%%%%%%%%%%%%%%%%%%%%%%%%%%%%%%%%%%%%%%%%%%%%

\section{Dataset}
\label{sec:dataset and implementation}
\subsection{Datasets}
We evaluate the effectiveness of our eMoE-Tracker through extensive experiments on two visible-event benchmark datasets: VisEvent~\cite{wang2023visevent} and COESOT~\cite{tang2022revisiting}.

\textbf{VisEvent.} VisEvent dataset contains 820 video sequence pairs including 37,128 RGB frames in total, and the minimum, maximum, and average frame lengths are 18, 6246, and 450 frames, respectively. The frame rate of RGB videos is around 25 FPS. The training subset contains 500 video sequences while the testing subset contains 320 video sequences. In the VisEvent dataset, there are 17 attributes defined, reflecting the scenarios under different lighting conditions, such as LI (Low Illumination), OE (Over Exposure), and IV (Illumination Variation).

\textbf{COESOT.} COESOT dataset is the largest benchmark dataset for RGB-event single object tracking. It comprises 1354 aligned video sequences captured by a DAVIS346 event camera, and the training subset contains 827 videos and the testing subset contains 527 videos, respectively. Similar to the VisEvent dataset, there are also 17 attributes annotated to help evaluate the performance of trackers under diverse scenarios.

\textbf{Attributes Selection.} For the four selected attributes, \ie, illumination variation, motion blur, scale variance and occlusion, on the one hand, they are categorized based on the 17 attributes in the testing sets, on the other hand, they can be based summarized on the observation as some samples shown in Fig.~\ref{sample}.

\textbf{Attributes Annotation.} In our eMoE-Tracker, we manually annotate the video sequences into a 4-digit vector according to environmental conditions. In particular, a video sequence is labeled as [illumination variation, motion blur, scale variance, occlusion] = $[1, 1, 0, 0]$, which means that the environmental condition in this video contains illumination variation and motion blue while without scale variance and occlusion. All the video sequences are labeled in this manner according to the RGB ones.

\section{Details of Experiments}
\label{sec.experiments}
\subsection{Implementation}
The eMoE-Tracker is trained on 1 NVIDIA A800 GPU with Pytorch implementation. The frozen ViT backbone structure is the same as the one in ViPT~\cite{zhu2023visual} and we pre-train the backbone ViT from scratch. We introduce four expert branches in this work, and they are with respect to illumination variation, motion blur, scale variance and occlusion. All four experts have the same structure and are initialized following a truncated normal distribution.

\subsection{Ablation Studies}
\textbf{The number of experts.} We represent the tracking results with the experts' number of 1,2,3,4 in TABLE~\ref{tab:experts}. When the number of experts is less than four, the attributes are randomly selected from the four pre-defined attributes and there should be more combinations to evaluate. Moreover, due to the manual annotations in existing settings, we are unable to evaluate the model with four more experts. This is the limitation of manual annotation for extension.

\textbf{Model complexity.} We show four trackers' trainable parameters to compare their complexity. Our eMoE-Tracker is with one-stream structure while others are two-stream ones. It should be clarified that a one-stream tracker with transformer structure is supposed to have more trainable parameters, \eg, CEUTrack~\cite{tang2022revisiting} has 96MB trainable parameters. However, our model eMoE-Tracker can gain better performance with less trainable parameters in one-stream trackers, resulting from the frozen ViT backbone reducing the trainable parameters.

%\subsection{Visual Prompt Learning}

\begin{figure}[t!]
\captionsetup{font=small}
\begin{center}
\includegraphics[width=\linewidth]{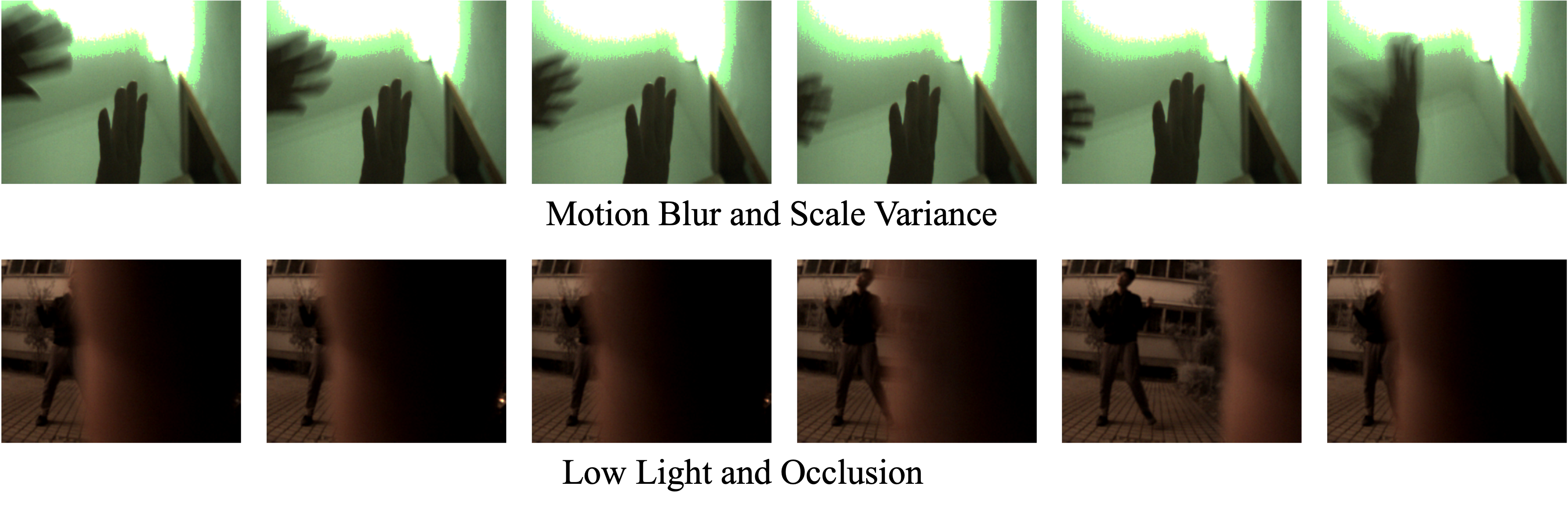}
\end{center}
   \caption{RGB image samples from COESOT dataset.}
\label{layer_attention}
\end{figure}
\section{Additional Evaluation Results}
\label{sample}
\subsection{Visualization}
As shown in Fig.~\ref{visualization}, we illustrate the attention maps from the backbone network and our eMoE-Tracker, which is from the last layer of the ViT encoder. In this part, we show more attention map results from layer 7 to 12 in Fig.~\ref{layer_attention}. From the attention maps from layer 7th to layer 12th, it is obvious that the responses from our eMoE-Tracker are clearer no matter in shallow or deep layers.

\begin{figure}[t!]
\captionsetup{font=small}
\begin{center}
\includegraphics[width=\linewidth]{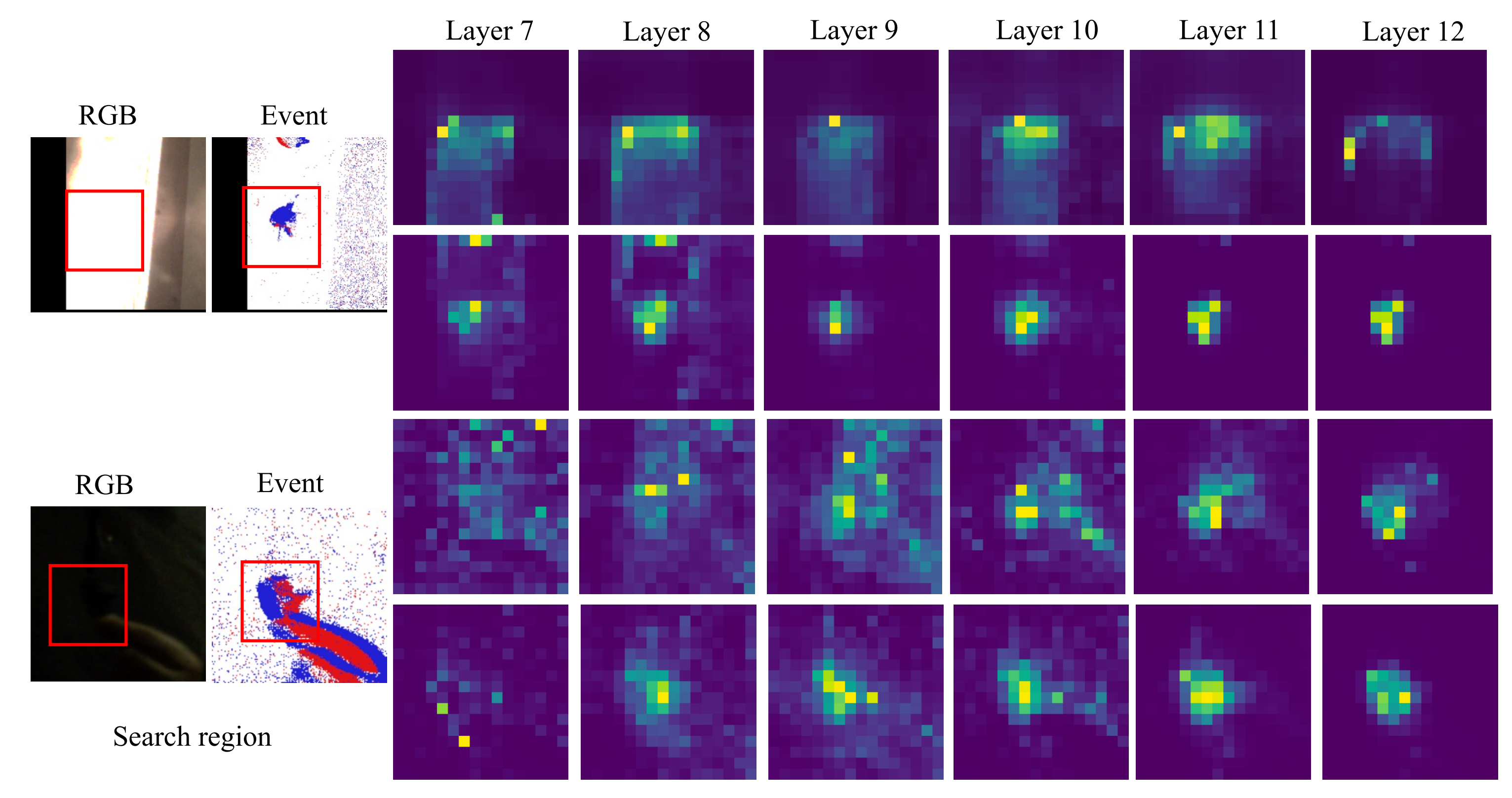}
\end{center}
\vspace{-5pt}
   \caption{The visualization of the attention maps of the 7th~12th layer from the backbone network and eMoE-Tracker. The left two columns are RGB and event search regions and the rest are the attention maps from two networks. The upper row is from backbone network while the lower one is from eMoE-Tracker.}
\label{layer_attention}
\end{figure}

\begin{table}[t!]
\centering
\caption{Testing Result in FELT-SOT dataset.}
\vspace{-5pt}
\fontsize{5}{6}\selectfont  
\resizebox{0.8\linewidth}{!}{
\begin{tabular}{c|lll}
\toprule
Dataset&SR&PR&NPR \\
\hline
FELT-SOT~\cite{wang2024FELTSOT}&65.1&72.9&73.6\\
\bottomrule
\end{tabular}}
\label{tab:real-world}
\end{table}

\subsection{Attributes Performance}
\begin{figure*}[h]
\captionsetup{justification=centerlast, font=small}
\begin{center}
\includegraphics[width=\linewidth]{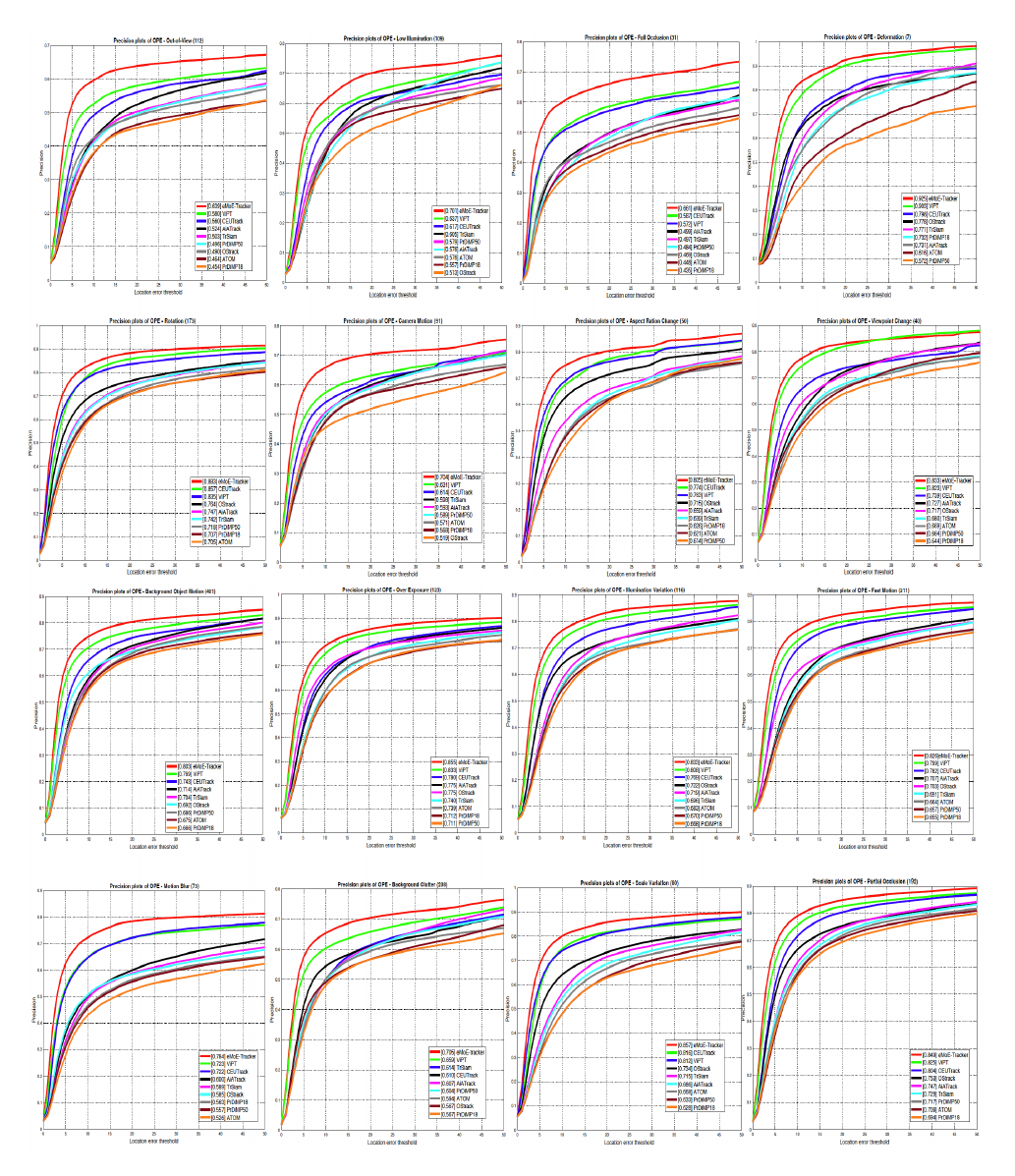}
\end{center}
\vspace{-5pt}
   \caption{The precision diagram of curves on attributes performance on COESOT dataset. We illustrate performance on 17 attributes, including Background Object Motion, Over-exposure, Illumination Variation, Rotation, Fast Motion, Aspect Ratio Change, Motion Blur, Background Clutter, Scale Variation, Viewpoint Change, Partial Occlusion, Out-of-View, Low Illumination, Full Occlusion, Deformation, and Camera Motion.}
\label{coesot_plot}
\vspace{-12pt}
\end{figure*}

\begin{figure*}[h]
\captionsetup{justification=centering, font=small}
\begin{center}
\includegraphics[width=\linewidth]{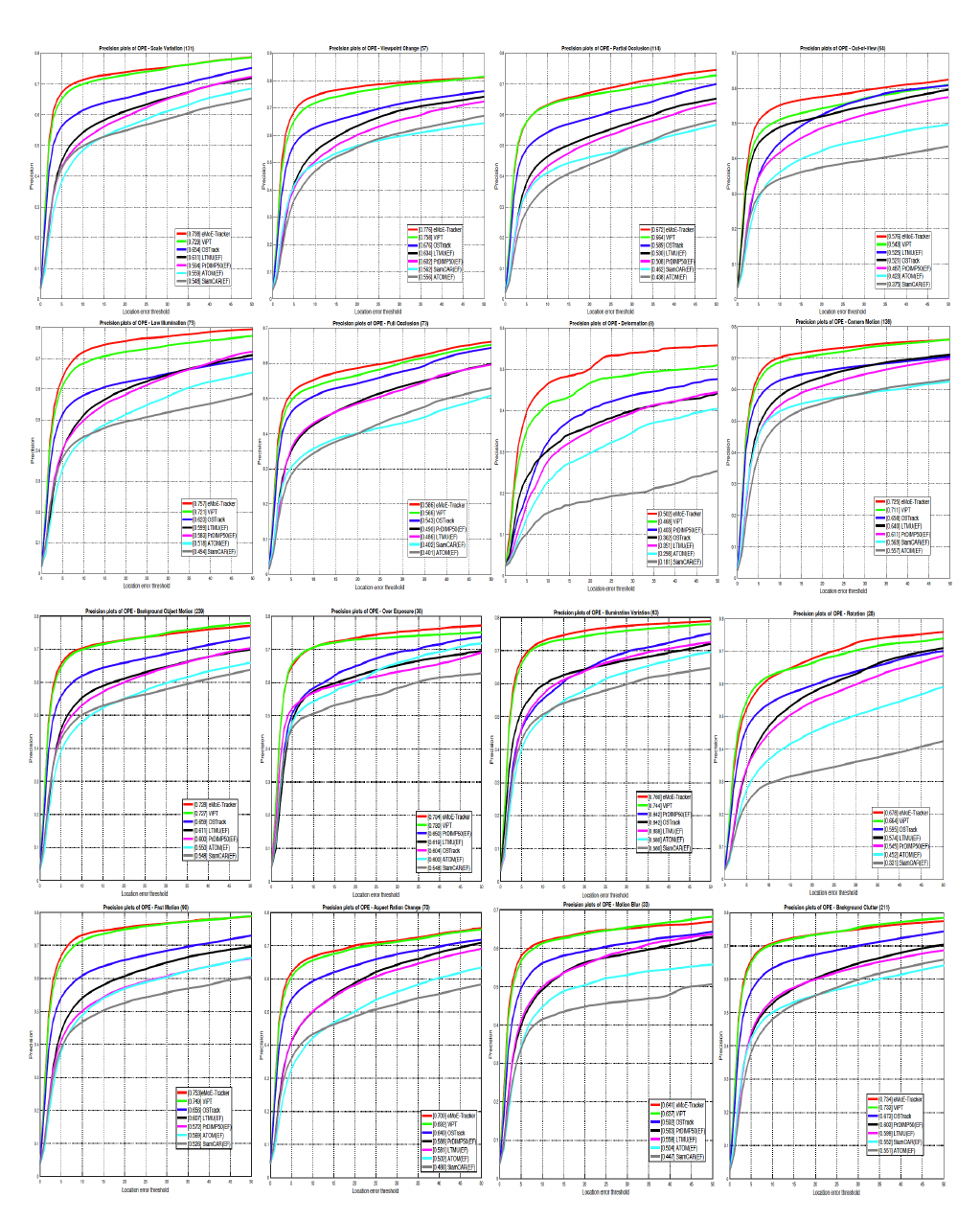}
\end{center}
\vspace{-5pt}
   \caption{The precision diagram of curves on attributes performance on VisEvent dataset. We illustrate performance on 17 attributes, including Background Object Motion, Over-exposure, Illumination Variation, Rotation, Fast Motion, Aspect Ratio Change, Motion Blur, Background Clutter, Scale Variation, Viewpoint Change, Partial Occlusion, Out-of-View, Low Illumination, Full Occlusion, Deformation, and Camera Motion.}
\label{visevent_plot}
\end{figure*}

\textbf{Attention Map.} Since the VisEvent and COESOT datasets all provide 17 attributes for tracking performance evaluation, we can leverage the Matlab toolkit to plot the diagram of curves about the attributes performance comparison. We represent the diagram of curves on 16 attributes except for the no motion scenario in Fig.~\ref{visevent_plot} and Fig.~\ref{coesot_plot}, which show the superior performance under all the environmental conditions compared to other SOTA priors.

\begin{figure}[t!]
\captionsetup{font=small}
\begin{center}
\includegraphics[width=\linewidth]{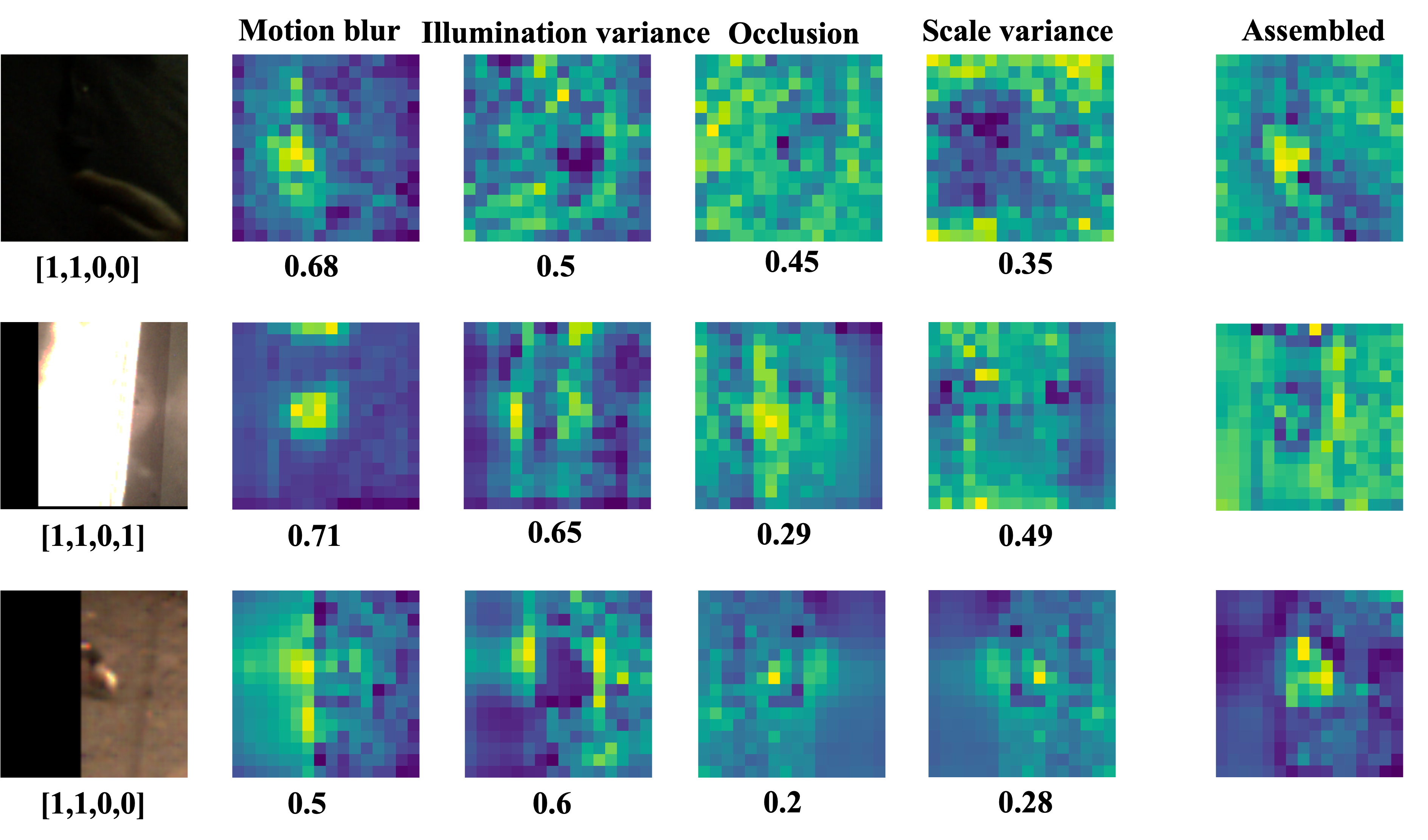}
\end{center}
\vspace{-5pt}
   \caption{The visualization of the feature maps and assembling weights for four environmental experts branches under specific scenarios. The 4-digit vector is the attributes label and the following numbers are the corresponding learnable weights for attribute-specific features. The last column denotes the assembled features.}
\label{feature}
\end{figure}

\textbf{Feature Map.} To validate that the disentangled environmental experts can learn discriminative attribute-specific features and assemble them in proper weights, we illustrate features and weights from four environmental experts in Fig.~\ref{feature}.

\subsection{Real-World Testing}
For real-world testing, we further validate our method on another RGB-event dataset, namely FELT-SOT~\cite{wang2024FELTSOT}, to demonstrate the practicality. The FELT-SOT~\cite{wang2024FELTSOT} dataset is a new long-term and large-scale frame-event single object tracking dataset. The testing result is reported in Table.~\ref{tab:real-world}.

\subsection{Inference Speed}
In object tracking tasks, Frames Per Second (FPS) is used to evaluate the inference speed of tracking methods. For deploying the tracking model on real machines, the inference speed should be greater than 25 FPS. We calculate the inference speed of our model and compare it with existing priors. The result is illustrated in Table.~\ref{tab:inference-speed}.

\vspace{-5pt}
\begin{table}[t!]
\centering
\caption{Comparison of inference speed between our eMoE-Tracker and other existing priors.}
\fontsize{7}{9}\selectfont  
\resizebox{\linewidth}{!}{
\begin{tabular}{c|llll}
\toprule
Trackers&OSTrack~\cite{ye2022joint}&ViPT~\cite{zhu2023visual}&CEUTrack~\cite{tang2022revisiting}&eMoE-Tracker \\
\hline
Inference speed&93.1&49&75&46 \\
\hline
\end{tabular}}
\vspace{-15pt}
\label{tab:inference-speed}
\end{table}

\section{Conclusion}
\label{sec:conclusion}
In supplementary material, we provide more details on datasets, experiments, and performance results for the evaluation. All the results show the effectiveness of our proposed eMoE-Tracker for RGB-event tracking. In the future, we are expected to extend the environmental expert branches dynamically and design the agent to detect the environmental conditions while not in a manual manner.

\end{document}